\documentclass{article}

\usepackage[final]{neurips_2022}

\usepackage[utf8]{inputenc} %
\usepackage[T1]{fontenc}    %
\usepackage{hyperref}       %
\usepackage{url}            %
\usepackage{booktabs}       %
\usepackage{amsfonts}       %
\usepackage{nicefrac}       %
\usepackage{microtype}      %
\usepackage{xcolor}         %
\usepackage{graphicx}
\usepackage{wrapfig}
\usepackage{capt-of}
\usepackage{float}
\usepackage{bbm}
\usepackage{listings}
\usepackage{longtable}
\usepackage{adjustbox}
\usepackage{wrapfig}

\usepackage{amsmath}
\usepackage{amssymb}

\usepackage{todonotes}

\usepackage[inline]{enumitem}
\usepackage{booktabs}
\usepackage{adjustbox}
\usepackage{multirow}
\newcommand{\prm}{{\mathsf{Prop Match}}}
\newcommand{\cls}{{\mathsf{Closeness}}}
\newcommand*\samethanks[1][\value{footnote}]{\footnotemark[#1]}

\title{LIC-GAN: Language Information Conditioned Graph Generative GAN Model}

\author{
    Robert Lo\thanks{~~Equal contribution.} \quad
    Arnhav Datar\samethanks \quad
    Abishek Sridhar\samethanks \quad\\
    School of Computer Science, Carnegie Mellon University\quad\quad \\
  {\tt \{chifanl,adatar,abisheks\}@cs.cmu.edu} \\
}
\lstdefinestyle{mystyle}{
    commentstyle=\color{codegreen},
    keywordstyle=\color{magenta},
    numberstyle=\tiny\color{gray},
    stringstyle=\color{codepurple},
    basicstyle=\ttfamily\footnotesize,
    breakatwhitespace=false,         
    breaklines=true,                 
    captionpos=b,                    
    keepspaces=true,                 
    numbers=left,                    
    numbersep=5pt,                  
    showspaces=false,                
    showstringspaces=false,
    showtabs=false,                  
    tabsize=2
}

\usepackage{xspace}
\newcommand{\bert}{\textsc{BERT}\xspace}
\newcommand{\roberta}{\textsc{RoBERTa}\xspace}
\newcommand{\chatgpt}{\textsc{GPT-3.5}\xspace}
\newcommand{\licgan}{\textsc{LIC-GAN}\xspace}
\newcommand{\molgan}{\textsc{MolGAN}\xspace}

\newcommand{\node}{\textsc{Node}\xspace}
\newcommand{\edge}{\textsc{Edge}\xspace}
\newcommand{\mindeg}{\textsc{MinDeg}\xspace}
\newcommand{\maxdeg}{\textsc{MaxDeg}\xspace}
\newcommand{\diameter}{\textsc{Diam}\xspace}
\newcommand{\ccnum}{\textsc{CCNum}\xspace}
\newcommand{\cycle}{\textsc{Cycle}\xspace}

\begin{document}

\maketitle

\begin{abstract}
Deep generative models for Natural Language data offer a new angle on the problem of graph synthesis: by optimizing differentiable models that directly generate graphs, it is possible to side-step expensive search procedures in the discrete and vast space of possible graphs. We introduce \licgan, an implicit, likelihood-free generative model for small graphs that circumvents the need for expensive graph matching procedures. Our method takes as input a natural language query and using a combination of language modelling and Generative Adversarial Networks (GANs) and returns a graph that closely matches the description of the query. We combine our approach with a reward network to further enhance the graph generation with desired properties. Our experiments, show that \licgan does well on metrics such as $\prm$ and $\cls$ getting scores of $0.36$ and $0.48$. We also show that \licgan performs as good as \chatgpt, with \chatgpt getting scores of $0.40$ and $0.42$. We also conduct a few experiments to demonstrate the robustness of our method, while also highlighting a few interesting caveats of the model.

\end{abstract}

\section{Introduction}

We work on building natural language conditional graph generative models. Current graph generation literature mostly focuses on unconditional generation of molecules and proteins, while conditional generation is limited to deterministic graph generation and simple scene graphs. Although there are currently limited applications where a graph needs to be sampled from a distribution on a small scale, at a bigger scale there are applications like city planning, latent sub-graph identification, task assignments, and approximate shortest path identification. If such a model can be successfully built and trained, it should also be able to approximate a deterministic output (scene graph, for instance) or categorical distribution (instance segmentation for an image) over the space of graphs. As a first step towards addressing this problem, we propose LIC-GAN: a language conditioned GAN model inspired by the \molgan architecture. We plan to evaluate the effectiveness of our proposed method on a random graph dataset we create.  A potential application of the setting we adopt could be the generation of graphical test cases given natural language input. %

\subsection{Related Works}
 
With the advent of graph neural networks, there has recently been a lot of work in Graph representation learning~\citep{bronstein17, leskovec2017, khoshraftar2022survey}, however not much in the field of graph generation. We can categorize prior works into two categories: unconditional and conditional generation, which is discussed below.

\begin{figure}
    \centering
    \begin{minipage}{.6\textwidth}
        \centering
        \includegraphics[width=\linewidth]{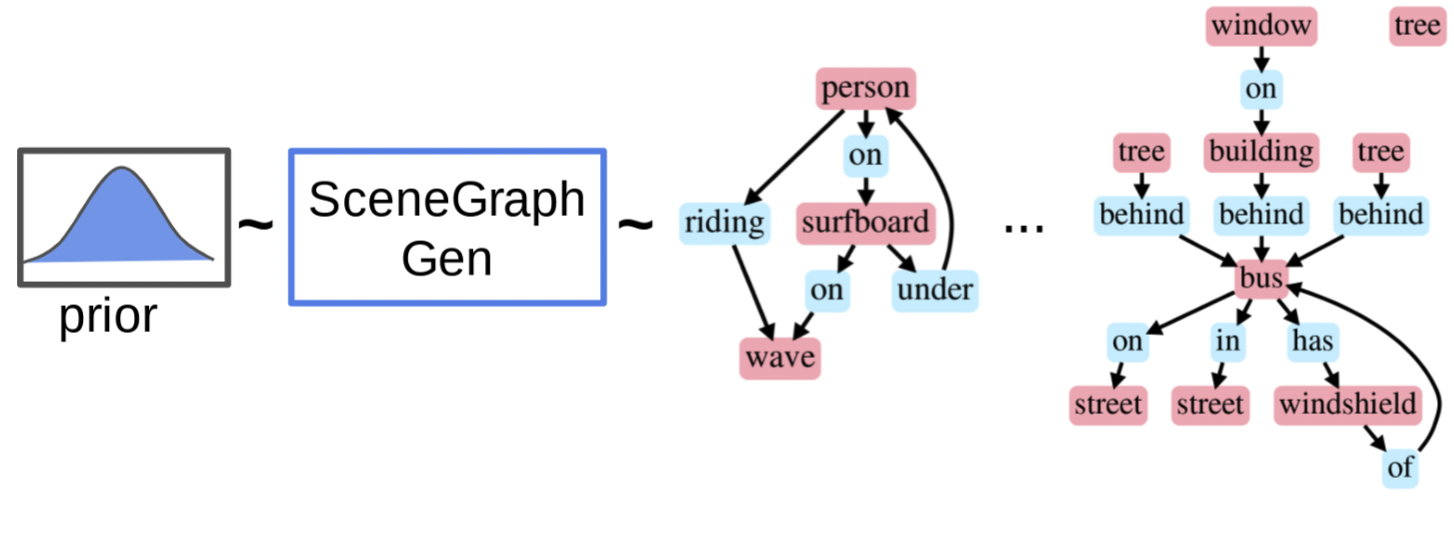}
        \caption{Unconditional Scene Graph Generation}
        \label{fig:unconditional}
    \end{minipage}%
    \begin{minipage}{0.4\textwidth}
        \centering
        \includegraphics[width=0.95\linewidth]{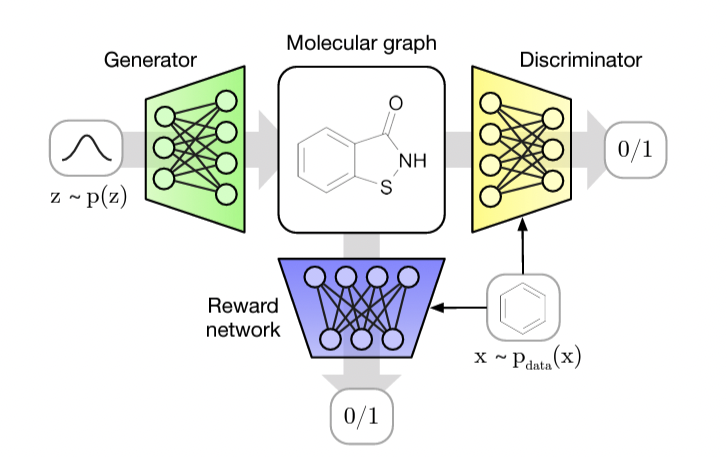}
        \caption{\molgan model schematics}
        \label{fig:molgan_model}
    \end{minipage}
\end{figure}

Unconditional graph generation is the task of generating graphs without any pre-specified constraints or conditions. These methods can be deterministic or can instead learn to sample from a distribution. This task has numerous applications in various fields such as social network analysis, chemical structure design, and image processing. In recent years, deep learning-based models have shown promising results in generating unconditional graphs~\citep{garg2021unconditional, wang_unconditional, kar2019metasim}. Figure \ref{fig:unconditional} shows the schematics for an unconditional scene graph generation model.

Discovering chemical compounds that possess specific characteristics can be a demanding endeavor with significant practical uses, such as creating novel pharmaceuticals from scratch. Likelihood-based methods~\citep{li2018learning, simonovsky2018graphvae} for generating molecular graphs necessitate either a predetermined (or randomly selected) sequential depiction of the graph or a costly graph matching process to determine the probability of a generated molecule. This is because assessing the likelihood of all feasible node orderings is impractical, even for small-sized graphs. One of the most significant and recent work in this area has been in correspondence to \molgan~\citep{de2018molgan}. \molgan is an implicit, likelihood free generative model for small molecular graphs that circumvents the need for expensive graph matching procedures or node ordering heuristics of previous likelihood-based methods. The schematic of \molgan can be found in Figure \ref{fig:molgan_model}. Some other notable works on chemical compound discovery are~\cite{Gomez-Bombarelli2018, pmlr-v70-kusner17a, dai2018syntax}.

On the contrary, most methods in conditional generation have been on producing a deterministic output. Scene graph generation with image and/or caption supervision is a particularly interesting research topic in this domain \citep{zhong2021learning}. There have also been works on road network extraction with image conditioning \citep{belli2019imageconditioned}. An open source work, called GraphGPT, which generates knowledge graphs from a given text using appropriate prompting on GPT models has also became popular.

Language Models such as \bert~\citep{devlin2018bert} and \roberta~\citep{liu2019roberta} have become one of the most widely used models in the field of NLP, and have been applied to a wide range of tasks including question answering~\citep{WIDAD2022379}, sentiment analysis~\citep{hoang-etal-2019-aspect, li2019exploiting}, and language translation~\citep{zhu2020incorporating, yang2020towards}. \chatgpt, a language model developed by OpenAI, has proven to be a valuable tool for multiple applications~\citep{biswas2023potential, sallam2023chatgpt}. However, its applicability in processing or generating graphical data is yet to be fully tested. In this report, we provide the first test of \chatgpt's skills in this regard. Rewards have been highly used in Reinforcement Learning. Recently, they have been incorporated in GANs~\citep{zheng2021reward, 9849003, Xia_Zhou_Shi_Lu_Huang_2020} for stronger generalization and robustness.

\section{Proposed Methods}

In this section, we briefly describe the two methods we employed for graph generation. We initially talk about the \chatgpt baseline, where we use prompt the \chatgpt model to generate graphs based on a description. We then talk about our main contribution of the report: LIC-GAN. We briefly talk about its architecture and some provide some details as to how it was trained. 

\begin{figure}
\begin{minipage}[b]{1.0\textwidth}
    \centering
    \includegraphics[width=0.8\textwidth]{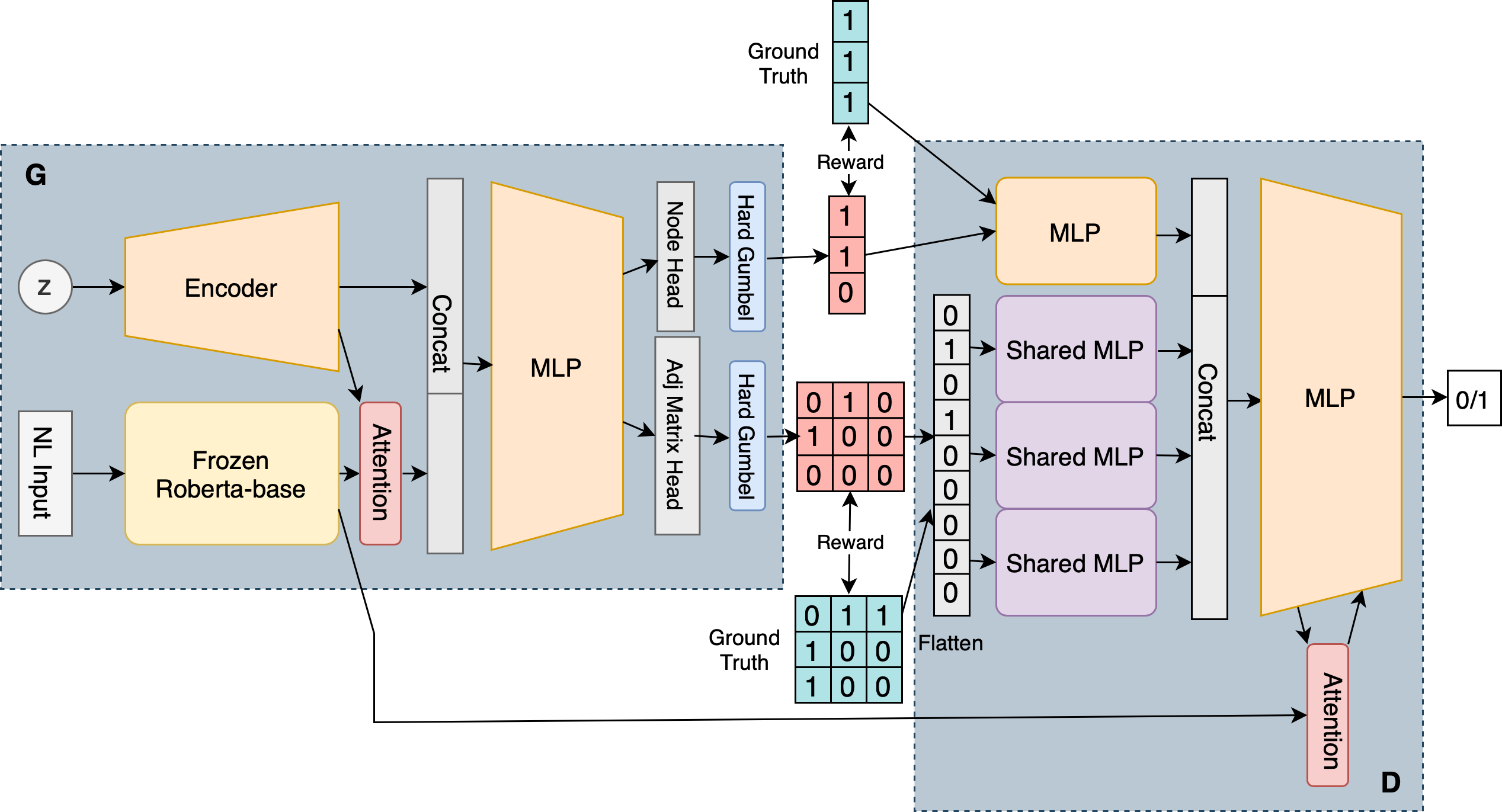}
    \caption{Proposed architecture for natural language conditioned GAN for graph generation}
    \label{fig:nlp_gan}
\end{minipage}
\end{figure}

\subsection{\chatgpt Baseline}
\label{sec:chatgpt_method}

Since, there has been no prior work, that has specifically tried to solve the problem of Natural Language Conditioned Graph Generation, we proposed a baseline method using \chatgpt. 

Within the prompt given to \chatgpt, we define what each of the 5 additional proposed properties mean in the graph-generation context. Furthermore, we also give it some example inputs and outputs.  

\chatgpt is a chatbot provided by OpenAI, which supports dialogue-like text completion. We designed a simple prompt that ask \chatgpt to generate the adjacency matrix. Since the adjacency matrices it generates are sometimes not perfect, we apply a simple post-process that pad the matrix and make it symmetric. The prompt, configuration and the post-processing method is given in \ref{appendix:prompt}.

\subsection{LIC-GAN}
\label{sec:gan_method}

\subsubsection*{Model Architecture}
The architecture of the final proposed model is shown in Figure \ref{fig:nlp_gan}. The initial models we used for preliminary analysis and architectural decisions had the adjacency matrix alone predicted, from which we obtain the nodes as the rows with all zeros in the matrix. This was later replaced to have two separate prediction heads for the node and edges, which lead to better performance and supported isolated vertices. As opposed to \molgan's choice of graph convolutional network (GCN) for the discriminator, we adopt a a weight-shared fully connected network (FCN) to process each row of the adjacency matrix to enforce the symmetry of the node processing (considering the nature of task).

\subsubsection*{Training Details}

WGANs are known to minimize an approximation of the Wasserstein-1 distance. The Wasserstein distance between two distributions $p$ and $q$ can be written as 

$$
D_W [p||q] = \frac{1}{K} \sup_{||f||_L \leq K} \left[ \mathbb{E}_{x \sim p(x)} [f(x)] - \mathbb{E}_{x \sim q(x)} [f(x)] \right]
$$

where supremum is taken from the set of all $K$-Lipschitz functions. We use the WGAN-GP loss formulation used by \molgan~\citep{de2018molgan} as well for training the model, to improve stability. Similar to the \molgan model, we also retain the gradient penalty from \cite{gulrajani2017improved} scaled by factor $\lambda_{\text{gp}} = 5$. For incorporating additional signals to train the GAN, we provide a reward in the form of negative of mean squared loss over node and edge number match. This loss is included in the overall loss after scaling by a factor $\lambda_{\text{rew}}$. 

For the preliminary analysis, we use a two hidden-layer FCN to predict the number of nodes from the adjacency matrix in a differentiable manner. For the final model, this becomes unnecessary due to separate dedicated heads. 

\section{Results}

In this section we discuss the results of our model. We initially describe the synthetic datasets that we used in our experiments in section \ref{sec:dataset} and how we evaluate the results of the generative methods in section \ref{sec:eval}. The results of the methods mentioned in the previous section are shown in section \ref{sec:chatgpt_results} and \ref{sec:gan_results}. 

\subsection{Dataset}
\label{sec:dataset}
To better analyze and benchmark our model, we create a synthetic graph dataset with natural language description of the properties. We utilized NetworkX~\citep{NetworkX} to generate four different types of undirected simple graphs (which corresponds to four different graph generation methods).

We consider two datasets: a \textbf{simple} dataset and a \textbf{complex} dataset, for better understanding of how our models are performing. The simple dataset only consists of the number of nodes and the number of edges in the graph in the textual description. Whereas the complex dataset consists of the number of nodes and the number of edges as well as a random subset of the 5 properties listed here: 
\begin{enumerate*}[label=(\alph*)]
    \item \diameter: the diameter \item \cycle: whether there exists a cycle or not \item \maxdeg: maximum degree \item \mindeg: minimum degree \item \ccnum: the number of connected components.
\end{enumerate*}

To make the dataset more versatile, we shuffle the properties in the description (for the model to not skip the language inputs and learn simpler correlations) in an online fashion for training all our final baselines and methods. The summary statistics for both datasets can be found in the Appendix \ref{appendix:summary_statistics}. For a fair comparision between different methods, we generated seperate datasets - train, dev and test, where we use only the test dataset to evaluate the performance all the methods presented in this report. The size of train, dev and test set is $100,000$, $10,000$ and $500$.

\subsection{Evaluation}
\label{sec:eval}

We primarily consider two metrics closeness and property match. They are defined as follows
\begin{align*}
   \prm &= \frac{1}{N} \sum_{i=1}^N \left( \frac{\sum_{p \in \mathsf{P_i}} \mathbbm{1}\left[p(G_{\mathsf{predicted}}) = p(G_{\mathsf{true}})  \right]} {|P_i|} \right) \\
    \cls &= \frac{1}{N} \sum_{i=1}^N \left( \frac{\sum_{p \in \mathsf{P_i}} \exp \left( - \left( p(G_{\mathsf{predicted}}) - p(G_{\mathsf{true}}) \right)^2\right)}{|P_i|} \right)
\end{align*}
Here $N$ is the number of data points that we are evaluating and $P_i$ is the set of properties that are present in the description of the $i^{\mathsf{th}}$ datapoint. It is clear to see that the $\prm  \leq \cls$, and that $\prm$ is only high if there is a perfect match where as $\cls$ is a bit forgiving to graphs that have properties that are close enough to the description. We also note that both metrics are going to lie in the $[0,1]$ range.

\subsection{\chatgpt Results}
\label{sec:chatgpt_results}

A brief summary of the results of the method described in Section \ref{sec:chatgpt_method} can be found Table \ref{tab:chatgpt_results_combined}. The detailed results can be found in Tables \ref{tab:chatgpt_results_simple} and \ref{tab:chatgpt_result_complex} in the Appendix. As can be seen from the results as we make the input more complex (increase the value of $n$), the property match and the closeness both start falling for the simple and the complex dataset. As expected, we can also observe that the complex dataset has significantly lower values of node $\prm$ and edge $\prm$ as can be seen from Tables  \ref{tab:chatgpt_results_simple} and \ref{tab:chatgpt_result_complex}. Particularly we note that the node $\prm$ went from $0.969$ to $0.178$ for the simple dataset as we transitioned from $n \in [11, 25]$ to $n \in [26, 50]$. Similarly it dipped from $0.9$ to $0.139$ for the complex dataset. This seems to imply that this method is unlikely to scale for larger graphs. 

\begin{table}
    \centering
    \begin{tabular}{l|l|*{5}{c}}
    \toprule
    Dataset & Metric & Overall & $n \in [1,5]$ & $n \in [6,10]$ & $n \in [11,25]$ & $n \in [26,50]$ \\ 
    \midrule
    \midrule
        \multirow{2}{*}{Simple} & $\prm$  & 0.423 & 0.933 & 0.635 & 0.522 & 0.097 \\ 
        & $\cls$ & 0.468 & 0.958 & 0.697 & 0.542 & 0.154 \\ 
        \midrule
        \multirow{2}{*}{Complex} &  $\prm$  & 0.397 & 0.528 & 0.514 & 0.462 & 0.233 \\ 
        & $\cls$ & 0.415 & 0.610 & 0.564 & 0.458 & 0.237 \\ 
    \bottomrule
    \end{tabular}
    \caption{\chatgpt Performace on the Simple and Complex Dataset. We have split the dataset into 4 buckets based on the graph complexity and show the results across these buckets in the last 4 columns}
    \label{tab:chatgpt_results_combined}
\end{table}

\subsection{\licgan Preliminary Analysis}
\label{sec:gan_pre_results}
We initially trained our model for $50$ epochs on the simple dataset (without shuffling the properties) to validate our design choices, and hyperparameters. We describe some of the important design choices based on the results found in Table \ref{tab:gan_results_simple}.

\begin{itemize}
    \item \textbf{\textit{Hard Gumbel Sigmoid} vs \textit{Sigmoid} as output-activation}: We found that using \textit{Sigmoid} to convert logits to adjacency make the convergence of training harder. Therefore, we decide to use \textit{Hard Gumbel Sigmoid}, an extension of the Gumbel Softmax~\citep{jang2016categorical}%
    \item \textbf{\textit{Multi-Head Attention} vs \textit{\texttt{[CLS]} token embedding} in generator}: \textit{Multi-Head Attention} allows the generator to access the whole embedding of the description, as opposed to \textit{[CLS] token embedding} only allows the generator to access the first token embedding. While it is common to represent the whole input with \textit{[CLS] token embedding}, we found that in our case using \textit{Multi-Head Attention} gives us a more balanced $\prm$ performance.
    \item \textbf{\textit{\roberta} vs \textit{\bert} in generator}: We found that the performance for \bert and \roberta is similar on our dataset. However, we decide to use \roberta because it has a better pretrain performance which might be helpful when the datasett is contains more complex natural language description (the text description in our dataset contains only simple words such as ``graph'', ``node'' and ``edge'').
    \item \textbf{\textit{FCN (Fully Connected Network)} vs \textit{GCN (Graph Convolution Network)} in discriminator}: We decide to use FCN instead of GCN because the former gives us a better performance. 
    \item \textbf{\textit{Adjacency matrix only} vs \textit{Adjacency matrix and node numbers}}: We found that generating only adjacency matrix and use heuristic to get the number of nodes give us poor performance compared to \chatgpt. To mitigate this and to support isolated vertices (e.g. vertex with degree $0$), our generator has two output head: a adjacency matrix prediction head and a node prediction head (to indicate the active nodes in the output graph). 
\end{itemize}

We can see that even when we shuffle the positions of the nodes and edges in the description, our model gives a similar performance. Similarly when we use text in the description as opposed to numbers (e.g ``two'' instead of ``2'') we are still able to perform to a similar level. This seems to indicate that the language model is appropriately embedding the information in the description to the feature vector.

\begin{table}
    \centering
    \begin{tabular}{l|l|l|c|c|c|c}
    \toprule
    $\lambda_{\text{rew}}$ & LM & Other Notes & $\prm$ & $\cls$ & \node PM &  \edge PM \\
    \midrule
    \midrule
    0 & \bert & & 0.2444 & 0.3287 & 0.3607 & 0.1281 \\ 
    0.5 & \bert & & 0.2397 & 0.3251 & 0.3578 & 0.1215 \\ 
    0.5 & \bert & Text input & 0.2231 & 0.3235 & 0.3211 & 0.1251 \\ 
    \midrule
    0 & \roberta & & 0.2437 & 0.3276 & 0.3159 & 0.1235 \\ 
    0 & \roberta & \textit{[CLS]} tokens & 0.2668 & 0.3391 & 0.4587 & 0.0751 \\ 
    0.5 & \roberta & \textit{[CLS]} tokens & 0.2405 & 0.3220 & 0.3614 & 0.1196 \\ 
    0.5 & \roberta & gcns & 0.1928 & 0.2898 & 0.2676 & 0.1181 \\ 
    \midrule
    0.5 & \bert & Invert $n$ and $m$ & 0.2228 & 0.314 & 0.3397 & 0.1058 \\ 
    0.5 & \bert & Shuffle $n$ and $m$ & 0.2139 & 0.3135 & 0.3007 & 0.1271 \\ 
    0.5 & \roberta & \textit{[CLS]} + Invert $n$ and $m$ & 0.2241 & 0.3135 & 0.3595 & 0.0887 \\ 
    0.5 & \roberta & \textit{[CLS]} + Shuffle $n$ and $m$ & 0.2244 & 0.3155 & 0.3371 & 0.1117 \\ 
    \bottomrule    
    \end{tabular}
    \caption{Initial results of \licgan on the simple dataset. LM represents the Language model used on the text description. Node PM and Edge PM represent the property match obtained when we just consider $P_i = \{ \mathsf{Node}\}$ and $P_i = \{ \mathsf{Edge}\}$ respectively for all $i = 1$ to $N$. }
    \label{tab:gan_results_simple}
\end{table}

\subsection{\licgan Final Results}
\label{sec:gan_results}
\begin{figure}
\begin{minipage}[b]{0.48\textwidth}
    \centering
    \includegraphics[width=0.8\textwidth]{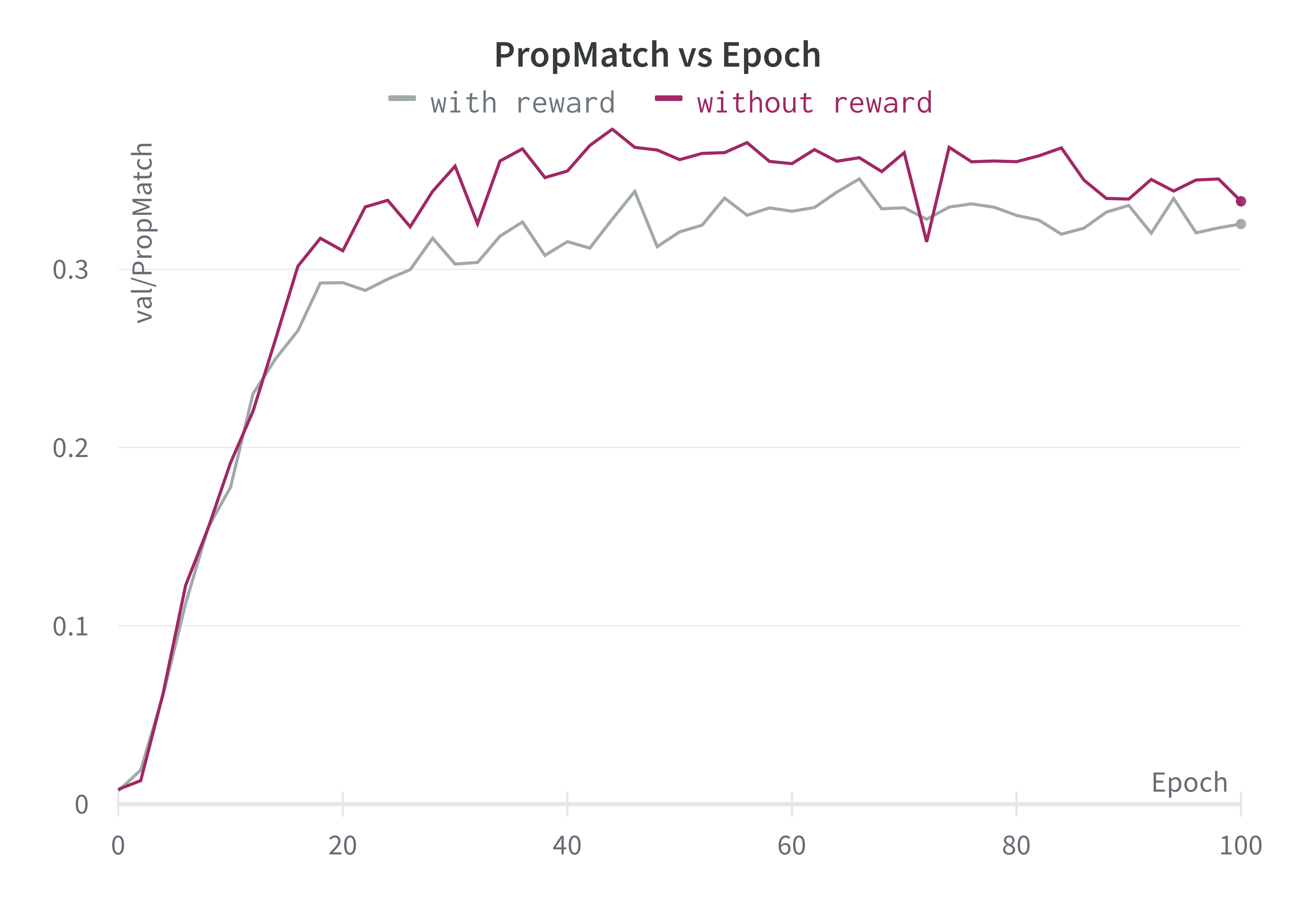}
    \caption{Property Match for final model on simple dataset with $\left(\lambda_{\text{rew}} = 0.5\right)$ and without reward}
    \label{fig:pmatch_simple}
\end{minipage}\hfill
\begin{minipage}[b]{0.48\textwidth}
    \centering
    \includegraphics[width=0.8\textwidth]{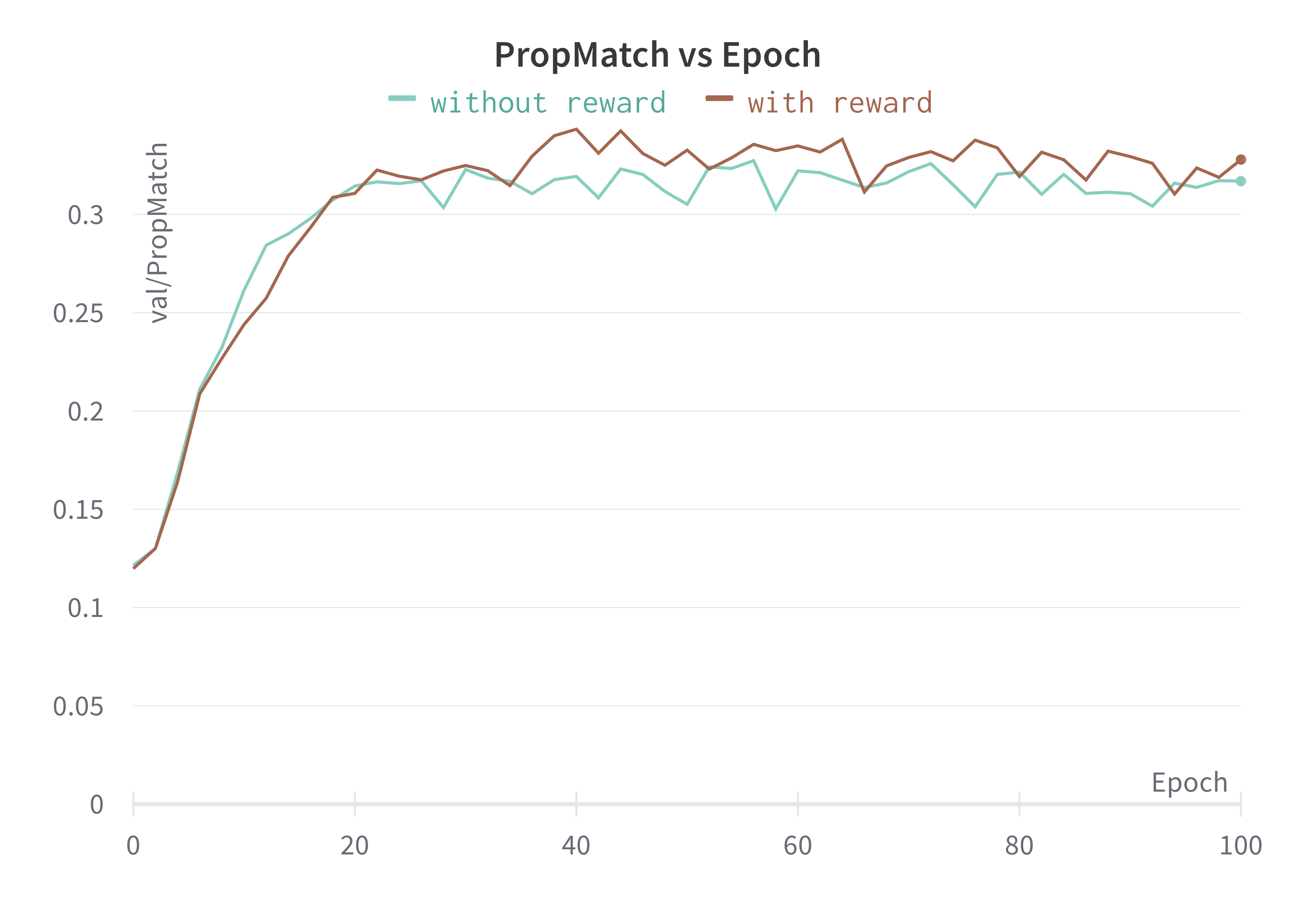}
    \caption{Property Match for final model on complex dataset with $\left(\lambda_{\text{rew}} = 0.5\right)$ and without reward}
    \label{fig:pmatch_complex}
\end{minipage}
\end{figure}

For our final results, we train the models for $100$ epochs with a constant $2\times 10^{-4}$ learning rate for the generator and discriminator and an \texttt{Adam} optimizer. The results can be found in Table \ref{tab:licgan_results_combined}, with a more detailed version in Tables \ref{tab:gan_results_simple_final} and \ref{tab:gan_result_complex} in the Appendix. While the reward helps for the complex dataset (especially for the node and edge match), it does not aid training on the simple dataset as seen in Figures \ref{fig:pmatch_simple} and  \ref{fig:pmatch_complex}.

It is observed that, while there is certainly a drop in performance as we increase $n$ for $\prm$ and $\cls$ for both the simple and the complex dataset, the drop is significantly lesser than the one for \chatgpt. This seems to imply that our \licgan model is likelier to scale for larger graphs unlike the \chatgpt model.

\begin{table}
    \centering
    \begin{tabular}{l|l|*{5}{c}}
    \toprule
    Dataset & Metric & Overall & $n \in [1,5]$ & $n \in [6,10]$ & $n \in [11,25]$ & $n \in [26,50]$ \\ 
    \midrule
    \midrule
        \multirow{2}{*}{Simple} & $\prm$  &  0.34 & 0.57 & 0.45 & 0.27 & 0.3 \\ 
        & $\cls$ &  0.44 & 0.67 & 0.54 & 0.36 & 0.39 \\ 
        \midrule
        \multirow{2}{*}{Complex} &  $\prm$  & 0.36 & 0.47 & 0.43 & 0.4 & 0.26 \\ 
        & $\cls$ & 0.48 & 0.62 & 0.56 & 0.5 & 0.39 \\ 
    \bottomrule
    \end{tabular}
    \caption{\licgan Performace on the Simple and Complex Dataset. We have split the dataset into 4 buckets based on the graph complexity and show the results across these buckets in the last 4 columns}
    \label{tab:licgan_results_combined}
\end{table}

\begin{figure}
    \centering
    \includegraphics[width = 0.9\textwidth]{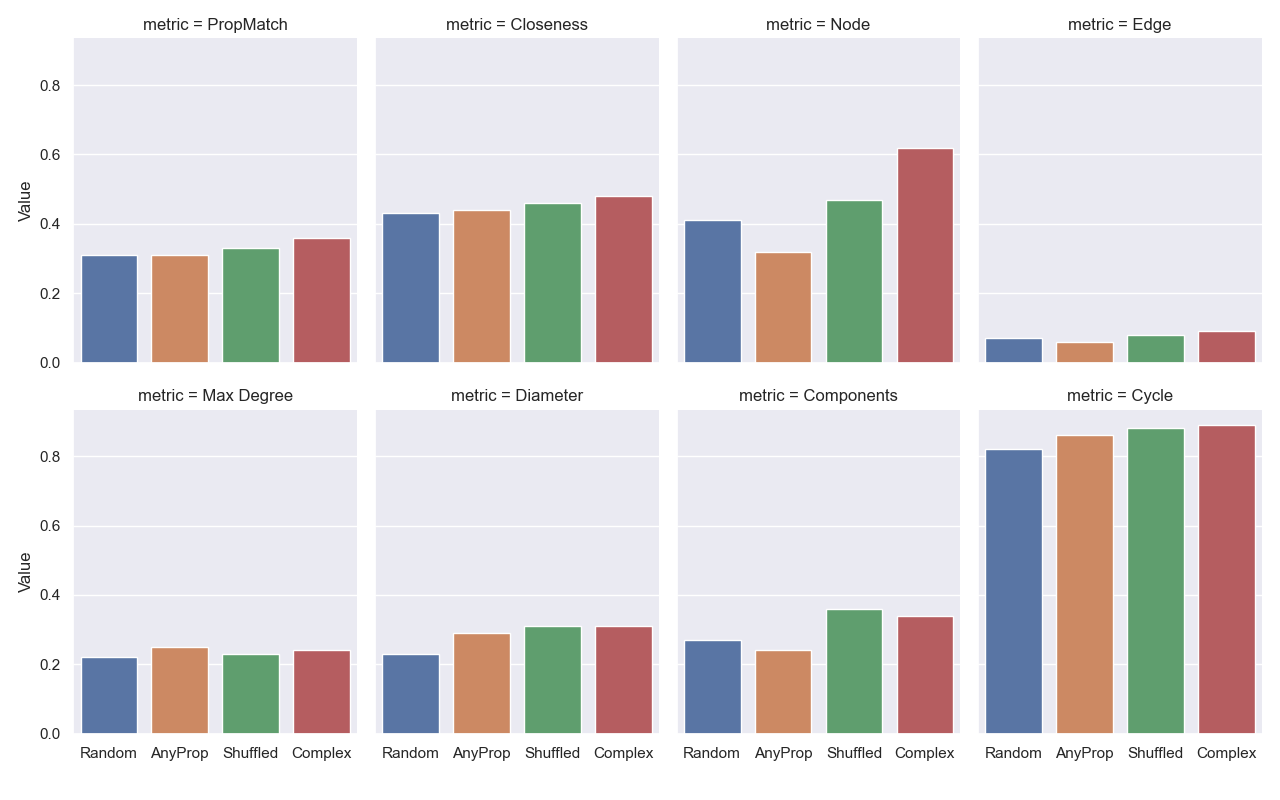}
    \caption{Plots for the performances for some metrics. The 4 experiments are 1) \textbf{Random}: We zero-shot test our trained model on descriptions which do not necessarily contain nodes/edges, 2) \textbf{AnyProp}: We train a new model that does not not necessarily contain nodes/edges in description in during train/test, 3) \textbf{Shuffled}: We zero-shot test descriptions just containing shuffled numeric values 4) \textbf{Complex}: LIC-GAN model's performance on the test set}
    \label{fig:hist_results}
\end{figure}

\section{Discussion and Analysis}

In this section, we run two experiments in order to do a ablation study on different parts of our generator e.g. language module and the graph generation module.  The \textit{shuffled numbers} experiments tries to demonstrate that the Language model is extracting crucial information from the textual description. The \textit{no node and edge} experiments try to show that the graph generation module can generate valid graph even without explicit constraint on graph number and edges. %

\subsection{The no nodes and edges experiments}

\begin{wrapfigure}{R}{7cm}
\centering
\includegraphics[width = 7cm]{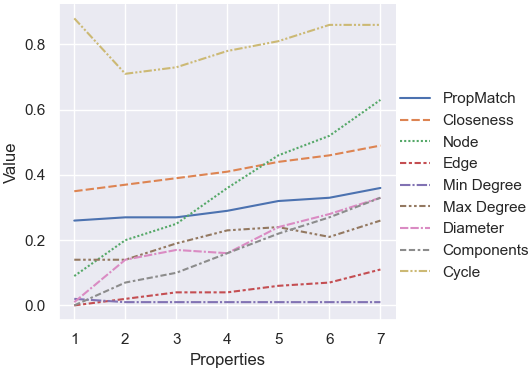}
\caption{The values of the various metrics as we increase the number of properties in the description during the 0-shot learning}
\label{fig:ValuesAndVariables}
\end{wrapfigure}

We test that the LIC-GAN model was not simply generating a lot of candidate graphs with matching $n$ and $m$ and randomly suggesting graphs that had properties that are close enough. To test our hypothesis, we zero-shot tested a variant of the complex dataset where we randomly choose $2-7$ properties (however, this time we do not compulsorily include nodes and edges in the description). As can be seen from Figure \ref{fig:hist_results} and from Table \ref{tab:random_gan}, we perform slightly worse in each metric. Nevertheless we are able to maintain a similar performance which seems to imply that our language model has truly learned the meaning of each of those properties. In fact, this variation of the complex dataset is what one might want to achieve in practice (without having to compulsorily provide any particular property). 

However, when we directly train the model on this version of the dataset, we achieve a worse performance compared to the zero-shot evaluation after training on the complex dataset. This validates the importance of implicit (and explicit) reward of preserving one or more easier properties in the description, that guides the training to a better solution. 

We also perform another experiment where we zero-shot test a randomly chosen set of properties and progressively increase the number of properties that we choose. The results can be found in Figure \ref{fig:ValuesAndVariables}. These results seem counter-intuitive as for a person if we are just provided a single property is very simple to come up with a graph that satisfies this property as opposed to a description with larger number of properties. However, this likely signals that our model can scale fairly well as we keep increasing the number of properties. The detailed table of results can be found in the Appendix in Table \ref{tab:zero-shot}.

\subsection{The shuffled numbers experiment}

We first show that our language model is able to extract information from textual description. We performed the following experiment: for every textual description such as ``10 nodes, 20 edges and minimum degree 1'' we just gave it the description ``20 1 10'', i.e. we removed the property names and shuffled the order of the input sequence. While we expected a significant decrease in performance as we made the task significantly harder, there is not a significant difference in the performance as can be seen from Figure \ref{fig:hist_results}. However, we do notice that the biggest performance drop has been in terms of \node's $\prm$. The detailed results for the same can be found in Table \ref{tab:random_gan} in the Appendix.

We hypothesize that this is might caused by the fact that \node and \edge  are always present (as opposed to the other properties) in the textual description and we have \edge $\geq$ \node int most cases (with the other properties all being less than \node). This allows the model to infer that the maximum number in the property vector is the required number of \edge, and the second largest number is the required number of \node. This explains why the sharpest drop in $\prm$ is observed for \node. The heatmap showing the probabilities of ordering of properties can be found in the Appendix in Figure \ref{fig:heatmap}. While we observe that there is a similar confusion between \maxdeg and \diameter, we believe that there isn't a significant performance drop here because both of these properties being present in the same description is an unlikely incident in our dataset (probability is around $30\%$). We also note that the most common ordering (\edge $\geq$ \node $\geq$ \maxdeg $\geq$ \diameter $\geq$ \ccnum $\geq$ \mindeg) appears around $\sim 60\%$ of the time in our dataset.

\section{Conclusion}

We studied the generation of Natural-Language Conditioned Graphs using GANs, however our work raises several open questions. While, we have primarily worked on generating the graph given the description, both models (\licgan and \chatgpt) knew that such a graph always existed. An interesting direction would be to have another model predict whether there exists even a single graph that satisfies all of the properties in the description. An example could be a description of the format ``A simple undirected graph with 3 nodes and 8 edges''. In conjunction with our model it can potentially generate a better language-conditioned graph generator. 

Another possible extension of our work is to make the dataset harder, such as generating directed graph and related properties, or generating weighted graph that satisfy some complex properties like value of min-cut. We can also make the undirected graph generation task harder by adding more advanced graph-theoretic properties (such as planarity, connectivity etc).

\subsection*{Access to Code}

All of the code for the project can be found \href{https://github.com/Arnhav-Datar/10708-Project}{here}. 

\newpage

\bibliography{sample}

\begin{thebibliography}{32}
\providecommand{\natexlab}[1]{#1}
\providecommand{\url}[1]{\texttt{#1}}
\expandafter\ifx\csname urlstyle\endcsname\relax
  \providecommand{\doi}[1]{doi: #1}\else
  \providecommand{\doi}{doi: \begingroup \urlstyle{rm}\Url}\fi

\bibitem[Belli and Kipf(2019)]{belli2019imageconditioned}
D.~Belli and T.~Kipf.
\newblock Image-conditioned graph generation for road network extraction, 2019.

\bibitem[Biswas(2023)]{biswas2023potential}
S.~S. Biswas.
\newblock Potential use of chat gpt in global warming.
\newblock \emph{Annals of biomedical engineering}, pages 1--2, 2023.

\bibitem[Bollob{\'a}s et~al.(2003)Bollob{\'a}s, Borgs, Chayes, and
  Riordan]{Bollobs2003DirectedSG}
B.~Bollob{\'a}s, C.~Borgs, J.~T. Chayes, and O.~Riordan.
\newblock Directed scale-free graphs.
\newblock In \emph{ACM-SIAM Symposium on Discrete Algorithms}, 2003.

\bibitem[Bronstein et~al.(2017)Bronstein, Bruna, LeCun, Szlam, and
  Vandergheynst]{bronstein17}
M.~M. Bronstein, J.~Bruna, Y.~LeCun, A.~Szlam, and P.~Vandergheynst.
\newblock Geometric deep learning: Going beyond euclidean data.
\newblock \emph{IEEE Signal Processing Magazine}, 34\penalty0 (4):\penalty0
  18--42, 2017.
\newblock \doi{10.1109/MSP.2017.2693418}.

\bibitem[Chen et~al.(2022)Chen, Yao, Wang, Sun, and Sheng]{9849003}
X.~Chen, L.~Yao, X.~Wang, A.~Sun, and Q.~Z. Sheng.
\newblock Generative adversarial reward learning for generalized behavior
  tendency inference.
\newblock \emph{IEEE Transactions on Knowledge and Data Engineering}, pages
  1--12, 2022.
\newblock \doi{10.1109/TKDE.2022.3186920}.

\bibitem[Dai et~al.(2018)Dai, Tian, Dai, Skiena, and Song]{dai2018syntax}
H.~Dai, Y.~Tian, B.~Dai, S.~Skiena, and L.~Song.
\newblock Syntax-directed variational autoencoder for molecule generation.
\newblock In \emph{Proceedings of the international conference on learning
  representations}, 2018.

\bibitem[De~Cao and Kipf(2018)]{de2018molgan}
N.~De~Cao and T.~Kipf.
\newblock {MolGAN: An implicit generative model for small molecular graphs}.
\newblock \emph{ICML 2018 workshop on Theoretical Foundations and Applications
  of Deep Generative Models}, 2018.

\bibitem[Devlin et~al.(2018)Devlin, Chang, Lee, and Toutanova]{devlin2018bert}
J.~Devlin, M.-W. Chang, K.~Lee, and K.~Toutanova.
\newblock Bert: Pre-training of deep bidirectional transformers for language
  understanding.
\newblock \emph{arXiv preprint arXiv:1810.04805}, 2018.

\bibitem[Erd\"os and R\'enyi(1959)]{Erdos:1959:pmd}
P.~Erd\"os and A.~R\'enyi.
\newblock On random graphs i.
\newblock \emph{Publicationes Mathematicae Debrecen}, 6:\penalty0 290--297,
  1959.

\bibitem[Garg et~al.(2021)Garg, Dhamo, Farshad, Musatian, Navab, and
  Tombari]{garg2021unconditional}
S.~Garg, H.~Dhamo, A.~Farshad, S.~Musatian, N.~Navab, and F.~Tombari.
\newblock Unconditional scene graph generation.
\newblock In \emph{Proceedings of the IEEE/CVF International Conference on
  Computer Vision}, pages 16362--16371, 2021.

\bibitem[G{\'o}mez-Bombarelli et~al.(2018)G{\'o}mez-Bombarelli, Wei, Duvenaud,
  Hern{\'a}ndez-Lobato, S{\'a}nchez-Lengeling, Sheberla, Aguilera-Iparraguirre,
  Hirzel, Adams, and Aspuru-Guzik]{Gomez-Bombarelli2018}
R.~G{\'o}mez-Bombarelli, J.~N. Wei, D.~Duvenaud, J.~M. Hern{\'a}ndez-Lobato,
  B.~S{\'a}nchez-Lengeling, D.~Sheberla, J.~Aguilera-Iparraguirre, T.~D.
  Hirzel, R.~P. Adams, and A.~Aspuru-Guzik.
\newblock Automatic chemical design using a data-driven continuous
  representation of molecules.
\newblock \emph{ACS Central Science}, 4\penalty0 (2):\penalty0 268--276, Feb
  2018.
\newblock ISSN 2374-7943.
\newblock \doi{10.1021/acscentsci.7b00572}.
\newblock URL \url{https://doi.org/10.1021/acscentsci.7b00572}.

\bibitem[Gulrajani et~al.(2017)Gulrajani, Ahmed, Arjovsky, Dumoulin, and
  Courville]{gulrajani2017improved}
I.~Gulrajani, F.~Ahmed, M.~Arjovsky, V.~Dumoulin, and A.~C. Courville.
\newblock Improved training of wasserstein gans.
\newblock \emph{Advances in neural information processing systems}, 30, 2017.

\bibitem[Hagberg et~al.(2008)Hagberg, Schult, and Swart]{NetworkX}
A.~A. Hagberg, D.~A. Schult, and P.~J. Swart.
\newblock Exploring network structure, dynamics, and function using networkx.
\newblock In G.~Varoquaux, T.~Vaught, and J.~Millman, editors,
  \emph{Proceedings of the 7th Python in Science Conference}, pages 11 -- 15,
  Pasadena, CA USA, 2008.

\bibitem[Hamilton et~al.(2017)Hamilton, Ying, and Leskovec]{leskovec2017}
W.~L. Hamilton, R.~Ying, and J.~Leskovec.
\newblock Representation learning on graphs: Methods and applications.
\newblock \emph{CoRR}, abs/1709.05584, 2017.
\newblock URL \url{http://arxiv.org/abs/1709.05584}.

\bibitem[Hoang et~al.(2019)Hoang, Bihorac, and Rouces]{hoang-etal-2019-aspect}
M.~Hoang, O.~A. Bihorac, and J.~Rouces.
\newblock Aspect-based sentiment analysis using {BERT}.
\newblock In \emph{Proceedings of the 22nd Nordic Conference on Computational
  Linguistics}, pages 187--196, Turku, Finland, Sept.{--}Oct. 2019.
  Link{\"o}ping University Electronic Press.
\newblock URL \url{https://aclanthology.org/W19-6120}.

\bibitem[Jang et~al.(2016)Jang, Gu, and Poole]{jang2016categorical}
E.~Jang, S.~Gu, and B.~Poole.
\newblock Categorical reparameterization with gumbel-softmax.
\newblock \emph{arXiv preprint arXiv:1611.01144}, 2016.

\bibitem[Kar et~al.(2019)Kar, Prakash, Liu, Cameracci, Yuan, Rusiniak, Acuna,
  Torralba, and Fidler]{kar2019metasim}
A.~Kar, A.~Prakash, M.-Y. Liu, E.~Cameracci, J.~Yuan, M.~Rusiniak, D.~Acuna,
  A.~Torralba, and S.~Fidler.
\newblock Meta-sim: Learning to generate synthetic datasets.
\newblock In \emph{ICCV}, 2019.

\bibitem[Khoshraftar and An(2022)]{khoshraftar2022survey}
S.~Khoshraftar and A.~An.
\newblock A survey on graph representation learning methods.
\newblock \emph{arXiv preprint arXiv:2204.01855}, 2022.

\bibitem[Kusner et~al.(2017)Kusner, Paige, and
  Hern{\'a}ndez-Lobato]{pmlr-v70-kusner17a}
M.~J. Kusner, B.~Paige, and J.~M. Hern{\'a}ndez-Lobato.
\newblock Grammar variational autoencoder.
\newblock In D.~Precup and Y.~W. Teh, editors, \emph{Proceedings of the 34th
  International Conference on Machine Learning}, volume~70 of \emph{Proceedings
  of Machine Learning Research}, pages 1945--1954. PMLR, 06--11 Aug 2017.
\newblock URL \url{https://proceedings.mlr.press/v70/kusner17a.html}.

\bibitem[Li et~al.(2019)Li, Bing, Zhang, and Lam]{li2019exploiting}
X.~Li, L.~Bing, W.~Zhang, and W.~Lam.
\newblock Exploiting bert for end-to-end aspect-based sentiment analysis.
\newblock \emph{arXiv preprint arXiv:1910.00883}, 2019.

\bibitem[Li et~al.(2018)Li, Vinyals, Dyer, Pascanu, and
  Battaglia]{li2018learning}
Y.~Li, O.~Vinyals, C.~Dyer, R.~Pascanu, and P.~Battaglia.
\newblock Learning deep generative models of graphs.
\newblock \emph{arXiv preprint arXiv:1803.03324}, 2018.

\bibitem[Liu et~al.(2019)Liu, Ott, Goyal, Du, Joshi, Chen, Levy, Lewis,
  Zettlemoyer, and Stoyanov]{liu2019roberta}
Y.~Liu, M.~Ott, N.~Goyal, J.~Du, M.~Joshi, D.~Chen, O.~Levy, M.~Lewis,
  L.~Zettlemoyer, and V.~Stoyanov.
\newblock Roberta: A robustly optimized bert pretraining approach.
\newblock \emph{arXiv preprint arXiv:1907.11692}, 2019.

\bibitem[Penrose(2003)]{penrose2003random}
M.~Penrose.
\newblock \emph{Random geometric graphs}, volume~5.
\newblock OUP Oxford, 2003.

\bibitem[Sallam et~al.(2023)Sallam, Salim, Barakat, and
  Al-Tammemi]{sallam2023chatgpt}
M.~Sallam, N.~Salim, M.~Barakat, and A.~Al-Tammemi.
\newblock Chatgpt applications in medical, dental, pharmacy, and public health
  education: A descriptive study highlighting the advantages and limitations.
\newblock \emph{Narra J}, 3\penalty0 (1):\penalty0 e103--e103, 2023.

\bibitem[Simonovsky and Komodakis(2018)]{simonovsky2018graphvae}
M.~Simonovsky and N.~Komodakis.
\newblock Graphvae: Towards generation of small graphs using variational
  autoencoders.
\newblock In \emph{Artificial Neural Networks and Machine Learning--ICANN 2018:
  27th International Conference on Artificial Neural Networks, Rhodes, Greece,
  October 4-7, 2018, Proceedings, Part I 27}, pages 412--422. Springer, 2018.

\bibitem[Wang et~al.(2019)Wang, Lin, Weissmann, Savva, Chang, and
  Ritchie]{wang_unconditional}
K.~Wang, Y.-A. Lin, B.~Weissmann, M.~Savva, A.~X. Chang, and D.~Ritchie.
\newblock Planit: Planning and instantiating indoor scenes with relation graph
  and spatial prior networks.
\newblock \emph{ACM Trans. Graph.}, 38\penalty0 (4), jul 2019.
\newblock ISSN 0730-0301.
\newblock \doi{10.1145/3306346.3322941}.
\newblock URL \url{https://doi.org/10.1145/3306346.3322941}.

\bibitem[Widad et~al.(2022)Widad, {El Habib}, and Ayoub]{WIDAD2022379}
A.~Widad, B.~L. {El Habib}, and E.~F. Ayoub.
\newblock Bert for question answering applied on covid-19.
\newblock \emph{Procedia Computer Science}, 198:\penalty0 379--384, 2022.
\newblock ISSN 1877-0509.
\newblock \doi{https://doi.org/10.1016/j.procs.2021.12.257}.
\newblock URL
  \url{https://www.sciencedirect.com/science/article/pii/S1877050921024960}.
\newblock 12th International Conference on Emerging Ubiquitous Systems and
  Pervasive Networks / 11th International Conference on Current and Future
  Trends of Information and Communication Technologies in Healthcare.

\bibitem[Xia et~al.(2020)Xia, Zhou, Shi, Lu, and
  Huang]{Xia_Zhou_Shi_Lu_Huang_2020}
Y.~Xia, J.~Zhou, Z.~Shi, C.~Lu, and H.~Huang.
\newblock Generative adversarial regularized mutual information policy gradient
  framework for automatic diagnosis.
\newblock \emph{Proceedings of the AAAI Conference on Artificial Intelligence},
  34\penalty0 (01):\penalty0 1062--1069, Apr. 2020.
\newblock \doi{10.1609/aaai.v34i01.5456}.
\newblock URL \url{https://ojs.aaai.org/index.php/AAAI/article/view/5456}.

\bibitem[Yang et~al.(2020)Yang, Wang, Zhou, Zhao, Zhang, Yu, and
  Li]{yang2020towards}
J.~Yang, M.~Wang, H.~Zhou, C.~Zhao, W.~Zhang, Y.~Yu, and L.~Li.
\newblock Towards making the most of bert in neural machine translation.
\newblock In \emph{Proceedings of the AAAI conference on artificial
  intelligence}, volume~34, pages 9378--9385, 2020.

\bibitem[Zheng et~al.(2021)Zheng, Yang, Parra-Ullauri, Garcia-Dominguez, and
  Bencomo]{zheng2021reward}
C.~Zheng, S.~Yang, J.~Parra-Ullauri, A.~Garcia-Dominguez, and N.~Bencomo.
\newblock Reward-reinforced reinforcement learning for multi-agent systems.
\newblock \emph{arXiv preprint arXiv:2103.12192}, 2021.

\bibitem[Zhong et~al.(2021)Zhong, Shi, Yang, Xu, and Li]{zhong2021learning}
Y.~Zhong, J.~Shi, J.~Yang, C.~Xu, and Y.~Li.
\newblock Learning to generate scene graph from natural language supervision,
  2021.

\bibitem[Zhu et~al.(2020)Zhu, Xia, Wu, He, Qin, Zhou, Li, and
  Liu]{zhu2020incorporating}
J.~Zhu, Y.~Xia, L.~Wu, D.~He, T.~Qin, W.~Zhou, H.~Li, and T.-Y. Liu.
\newblock Incorporating bert into neural machine translation.
\newblock \emph{arXiv preprint arXiv:2002.06823}, 2020.

\end{thebibliography}
\bibliographystyle{abbrvnat}

\newpage
\appendix

\section{Appendix}

\subsection{\chatgpt Prompts and Post-Processing}
\label{appendix:prompt}

\subsubsection*{Post-Processing method}
Suppose that given text description $T$, \chatgpt generate matrix $G\in \mathbb{R}^{n\times m}$. Note that $n$ is not necessary $m$, WLOG assume that $n\ge m$. The first step is to convert it to a square matrix by padding zeros, then make it symmetric by adding its transpose to itself. In other words, we pad $G$ to $G' \in \mathbb{R}^{n\times n}$ then update $G' = G'^T + G'$.

\subsubsection*{Configuration}

\begin{table}[H]
    \centering
    \begin{tabular}{l|*{5}{c}}
    \toprule
    Parameter & Value \\ 
    \midrule
    \midrule
    Endpoint & openai.ChatCompletion.create \\
    model & gpt-3.5-turbo \\
    temperature & 0 \\
    n & 1 \\
    max\_tokens & min(3500, $g^2+g+1$), where $g$ is the number of nodes \\
    top\_p & 1 \\
    frequency\_penalty & 0.0 \\
    presence\_penalty & 0.0 \\
    stop & \texttt{```end} \\
    \bottomrule
    \end{tabular}
    \caption{\chatgpt Configuration}
    \label{tab:chatgpt_config}
\end{table}

\subsubsection*{\chatgpt Prompts}
The prompt we use is given below, and is presented in a dialogue style because \chatgpt is a chatbot LLM. \texttt{TEXT\_DESCRIPTION} is replaced to the text description of graph that we want to generate now.

\begin{lstlisting}
[
    {
        'role': 'system',
        'content': "You are given a task to generate an adjacency matrix of an undirected graph with $n$ nodes and $m$ edges. The graph also need to satisfy several properties, which may includes (1) the min among nodes (2) max degree among nodes (3) number of connected components (4) diameter of graph (5) have cycle or not. Diameter of a graph is the maximum distance between any two nodes. A connected component is a subgraph in which any two nodes are connected to each other by paths, and which is connected to no other nodes in the supergraph. A cycle is a path of length at least 3 that starts and ends at the same node. \n\n    The graph may be unconnected, but every node has AT LEAST ONE EDGE. If you don't know how to generate a graph that satisfy all properties, you can output a graph that at least have the correct number of nodes and edges, and match other properties at your best effort. Output the adjacency matrix of the graph only without any other information."
    },
    {
        'role': 'system',
        'name': 'example_user',
        'content': 'Undirected graph with 9 nodes, 11 edges, 1 connected component, one or more cycle, min degree 1, max di
ameter 5, max degree 3.'
    },
    {
        'role': 'system',
        'name': 'example_assistant',
        'content': '```\n0 1 0 0 1 1 0 0 0\n1 0 0 0 0 0 0 1 1\n0 0 0 1 0 0 0 0 0\n0 0 1 0 0 0 1 0 1\n1 0 0 0 0 1 0 0 0\n1 0 0 0 1 0 0 1 0\n0 0 0 1 0 0 0 0 0\n0 1 0 0 0 1 0 0 1\n0 1 0 1 0 0 0 1 0\n```end'
    },
    {
        'role': 'user',
        'content': TEXT_DESCRIPTION
    }
]
\end{lstlisting}

\section{Dataset Generation Process and Summary Statistics}
\label{appendix:summary_statistics}

\subsection{Generation Process}
For each graph type, we first randomly choose one configuration among the four possible configurations. Each configuration has a node set $A_i$ associated with it, which represents a contiguous segment of numbers. Then, we sample node uniformly from set $A_i$, where $A_1=[5, 9], A_2=[10, 24], A_3=[25, 40], A_4=[40, 50]$.  Finally, we generate the graph with the sampled node and parameters. The configuration is given in table \ref{tab:graphgen-param}. We generate three set of data, namely \textit{train, dev, test}, with different amount of data. Train set has 100k samples, dev set has 10k and test set has 500.
\begin{table}[H]
\centering
\begin{tabular}{@{}l|cc@{}}
\toprule
Graph Type             & Count        & Parameters                               \\ \midrule \midrule
Scale Free Graph       & 30\% & n= $A_1, A_2, A_3, A_4$   \\
Erdos-Renyi Graph      & 30\% & (n,p)=($A_1$,0.3),($A_2$,0.2),($A_3$,0.1),($A_4,0.2$) \\
Random Geometric Graph & 30\% & (n,r)=($A_1$,0.5),($A_2$,0.4),($A_3$,0.2),($A_4$,0.2) \\
Uniform Random Tree    & 10\% & n=$A_1, A_2, A_3, A_4$ \\ 
\bottomrule
\end{tabular}

\caption{Methods, parameters and counts for four different graph type. Scale Free Graph \cite{Bollobs2003DirectedSG} is random graphs whose degree follow power law. Erdos-Renyi Graph \cite{Erdos:1959:pmd} is a random graph where each edge has a fixed probability of being created. Random Geometric Graph \cite{penrose2003random} is generated by first scatter points in a 2-d space, then connect points that are close enough. and Uniform Random Tree. $n$ represents number of nodes. $p$ represents the probability of edge creation. $r$ represents the distance threshold value. Please refer to the documentation of \href{https://networkx.org/documentation/stable/reference/generators.html}{NetworkX} for more details.}
\label{tab:graphgen-param}
\end{table}

\subsection{Summary Statistics}

For the following tables, \node represents number of nodes, \edge represents number of edges, \diameter represents the \href{https://en.wikipedia.org/wiki/Distance_(graph_theory)\#diameter}{diameter} of graph among all connected components, \ccnum represents the number of connected components, \maxdeg / \mindeg represents the maximum node degree / minimum node degree.

\subsubsection*{Train}
\begin{table}[H]
\centering
\begin{tabular}{@{}l|cccccc@{}}
\toprule
      & \node      & \edge      & \mindeg & \maxdeg & \diameter & \ccnum \\ \midrule \midrule
mean  & 22.346 & 37.906 & 1.196 & 8.617   & 5.481    & 1.227    \\
std   & 14.201 & 29.728 & 0.551         & 6.141   & 3.445    & 0.645    \\
min   & 5      & 3      & 0             & 1       & 2        & 1        \\
25\%  & 9      & 12     & 1             & 4       & 4        & 1        \\
50\%  & 19     & 32     & 1             & 7       & 4        & 1        \\
75\%  & 35     & 56     & 1             & 10      & 6        & 1        \\
max   & 50     & 189    & 7             & 42      & 30      & 8        \\ 
\bottomrule
\end{tabular}
\caption{Descriptive Statistics for train set. The size of train set is 100,000.}
\label{tab:train-graphgen-stat}
\end{table}

\subsubsection*{Dev}
\begin{table}[H]
\centering
\begin{tabular}{@{}l|cccccc@{}}
\toprule
      & \node      & \edge      & \mindeg & \maxdeg & \diameter & \ccnum \\ \midrule \midrule
mean  & 22.563 & 38.307 & 1.192 & 8.667   & 5.471    & 1.234    \\
std   & 14.286 & 30.014 & 0.556         & 6.224   & 3.445    & 0.649    \\
min   & 5      & 3      & 0             & 1       & 2        & 1        \\
25\%  & 9      & 12     & 1             & 4       & 4        & 1        \\
50\%  & 20     & 32     & 1             & 7       & 4        & 1        \\
75\%  & 35     & 56     & 1             & 10      & 6        & 1        \\
max   & 50     & 178    & 6             & 40      & 29       & 7        \\ 
\bottomrule
\end{tabular}
\caption{Descriptive Statistics for dev set. The size of train set is 10,000.}
\label{tab:dev-graphgen-stat}
\end{table}

\subsubsection*{Test}
\begin{table}[H]
\centering
\begin{tabular}{@{}l|cccccc@{}}
\toprule
      & \node      & \edge      & \mindeg & \maxdeg & \diameter & \ccnum \\ \midrule \midrule
mean  & 22.446 & 38.312 & 1.240 & 8.622   & 5.533    & 1.244    \\
std   & 14.330 & 30.418 & 0.622         & 5.894   & 3.776    & 0.720    \\
min   & 5      & 3      & 0             & 1       & 2        & 1        \\
25\%  & 9      & 11     & 1             & 4       & 3        & 1        \\
50\%  & 20     & 32     & 1             & 7       & 4        & 1        \\
75\%  & 35     & 57     & 1             & 10      & 6        & 1        \\
max   & 50     & 155    & 6             & 40      & 26       & 7        \\ 
\bottomrule
\end{tabular}
\caption{Descriptive Statistics for test set. The size of train set is 500.}
\label{tab:test-graphgen-stat}
\end{table}

\subsubsection*{Property Ordering}

\begin{figure}[H]
    \centering
    \includegraphics[width = 0.8\textwidth]{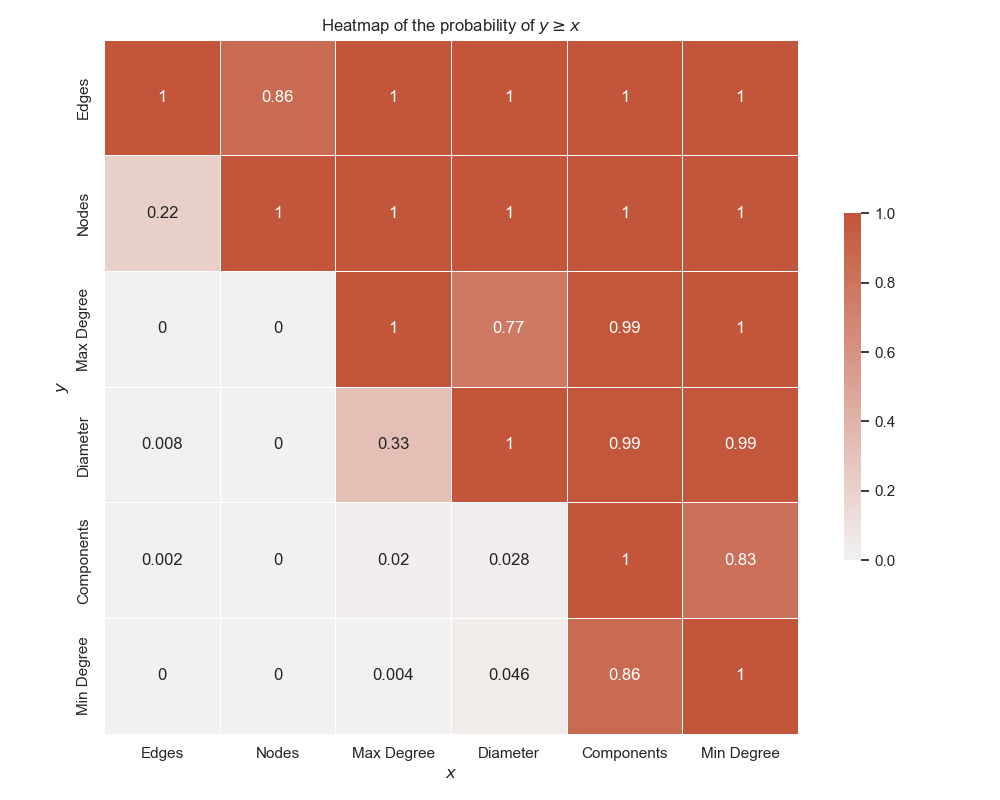}
    \caption{Heatmap showing the probabilities that $y \geq x$}
    \label{fig:heatmap}
\end{figure}

\section{Detailed Results for \chatgpt and LIC-GAN}

\begin{table}[H]
    \centering
    \begin{tabular}{l|*{5}{c}}
    \toprule
    & Overall & $n \in [1,5]$ & $n \in [6,10]$ & $n \in [11,25]$ & $n \in [25,50]$ \\ 
    \midrule
    \midrule
        $\prm$  & 0.423 & 0.933 & 0.635 & 0.522 & 0.097 \\ 
        $\cls$ & 0.468 & 0.958 & 0.697 & 0.542 & 0.154 \\ 
        \node $\prm$ & 0.690 & 1.000 & 0.985 & 0.969 & 0.178 \\ 
        \edge $\prm$ & 0.156 & 0.867 & 0.285 & 0.075 & 0.017 \\ 
    \bottomrule
    \end{tabular}
    \caption{\chatgpt Performace on the Simple Dataset}
    \label{tab:chatgpt_results_simple}
\end{table}

\begin{table}[H]
    \centering
    \begin{tabular}{l|*{5}{c}}
    \toprule
    & Overall & $n \in [1,5]$ & $n \in [6,10]$ & $n \in [11,25]$ & $n \in [25,50]$ \\
    \midrule
    \midrule
        $\prm$  & 0.397 & 0.528 & 0.514 & 0.462 & 0.233 \\ 
        $\cls$ & 0.415 & 0.610 & 0.564 & 0.458 & 0.237 \\ 
        \node $\prm$ & 0.626 & 0.867 & 0.908 & 0.900 & 0.139 \\ 
        \edge $\prm$ & 0.058 & 0.233 & 0.085 & 0.044 & 0.022 \\ 
        \mindeg $\prm$ & 0.266 & 0.250 & 0.212 & 0.367 & 0.218 \\ 
        \maxdeg $\prm$ & 0.153 & 0.524 & 0.344 & 0.070 & 0.031 \\ 
        \diameter $\prm$ & 0.130 & 0.353 & 0.319 & 0.094 & 0.000 \\ 
        \ccnum $\prm$ & 0.793 & 0.895 & 0.828 & 0.855 & 0.691 \\ 
        \cycle $\prm$ & 0.814 & 0.778 & 0.874 & 0.852 & 0.740 \\
    \bottomrule
    \end{tabular}
    \caption{\chatgpt Performace on the Complex Dataset. \diameter represents the maximum diameter of the graph. \ccnum represnts the number of connected components and \cycle represents whether a graph has a cycle. }
    \label{tab:chatgpt_result_complex}
\end{table}

\begin{table}[H]
    \centering
    \begin{tabular}{l|*{5}{c}}
    \toprule
    & Overall & $n \in [1,5]$ & $n \in [6,10]$ & $n \in [11,25]$ & $n \in [25,50]$ \\ 
    \midrule
    \midrule
        $\prm$  & 0.34 & 0.57 & 0.45 & 0.27 & 0.3 \\ 
        $\cls$ & 0.44 & 0.67 & 0.54 & 0.36 & 0.39 \\ 
        \node $\prm$ & 0.59 & 0.83 & 0.73 & 0.5 & 0.53 \\ 
        \edge $\prm$ & 0.1 & 0.3 & 0.16 & 0.04 & 0.08 \\
    \bottomrule
    \end{tabular}
    \caption{\licgan Performace on the Simple Dataset}
    \label{tab:gan_results_simple_final}
\end{table}

\begin{table}[H]
    \centering
    \begin{tabular}{l|*{5}{c}}
    \toprule
    & Overall & $n \in [1,5]$ & $n \in [6,10]$ & $n \in [11,25]$ & $n \in [25,50]$ \\
    \midrule
    \midrule
         $\prm$  & 0.36 & 0.47 & 0.43 & 0.4 & 0.26 \\ 
        $\cls$ & 0.48 & 0.62 & 0.56 & 0.5 & 0.39 \\ 
        \node $\prm$ & 0.62 & 0.9 & 0.78 & 0.62 & 0.46 \\ 
        \edge $\prm$ & 0.09 & 0.3 & 0.17 & 0.06 & 0.02 \\ 
        \mindeg $\prm$ & 0.02 & 0.04 & 0 & 0 & 0.05 \\ 
        \maxdeg $\prm$ & 0.24 & 0.33 & 0.3 & 0.26 & 0.16 \\ 
        \diameter $\prm$ & 0.31 & 0.47 & 0.4 & 0.35 & 0.19 \\ 
        \ccnum $\prm$ & 0.34 & 0.47 & 0.41 & 0.52 & 0.11 \\ 
        \cycle $\prm$ & 0.89 & 0.83 & 0.9 & 0.92 & 0.87 \\ 
    \bottomrule
    \end{tabular}
    \caption{\licgan Performace on the Complex Dataset. \diameter represents the maximum diameter of the graph. \ccnum represnts the number of connected components and \cycle represents whether a graph has a cycle. }
    \label{tab:gan_result_complex}
\end{table}

\begin{table}[H]
    \centering
    \begin{tabular}{cccccccccc}
    \toprule
    & PM  & $\cls$ & \node & \edge & \mindeg & \maxdeg & \diameter & \ccnum & \cycle \\  
    \midrule
    \midrule
        Shuffled & 0.33 & 0.46 & 0.47 & 0.08 & 0.02 & 0.23 & 0.31 & 0.36 & 0.88 \\
        Random & 0.31 & 0.43 & 0.41 & 0.07 & 0.01 & 0.22 & 0.23 & 0.27 & 0.82 \\ 
        Complex & 0.36 & 0.48 & 0.62 & 0.09 & 0.02 & 0.24 & 0.31 & 0.34 & 0.89 \\ 
    \bottomrule
    \end{tabular}
    \caption{PM represents $\prm$, for the other properties we represent their $\prm$ values. The random dataset was zero-shot tested on the trained model. The random dataset does not necessarily have the number of nodes and edges in the description, whereas the complex dataset always has the nodes and edges in the description. The shuffled dataset just has the shuffled numeric values with the textual description. }
    \label{tab:random_gan}
\end{table}

\begin{table}[H]
    \centering
    \begin{tabular}{cccccccccc}
    \toprule
    Props & PM  & $\cls$ & \node & \edge & \mindeg & \maxdeg & \diameter & \ccnum & \cycle \\  
    \midrule
    \midrule
        1 & 0.26 & 0.35 & 0.09 & 0 & 0.02 & 0.14 & 0.01 & 0 & 0.88 \\ 
        2 & 0.27 & 0.37 & 0.2 & 0.02 & 0.01 & 0.14 & 0.14 & 0.07 & 0.71 \\ 
        3 & 0.27 & 0.39 & 0.25 & 0.04 & 0.01 & 0.19 & 0.17 & 0.1 & 0.73 \\ 
        4 & 0.29 & 0.41 & 0.36 & 0.04 & 0.01 & 0.23 & 0.16 & 0.16 & 0.78 \\ 
        5 & 0.32 & 0.44 & 0.46 & 0.06 & 0.01 & 0.24 & 0.24 & 0.22 & 0.81 \\ 
        6 & 0.33 & 0.46 & 0.52 & 0.07 & 0.01 & 0.21 & 0.28 & 0.27 & 0.86 \\ 
        7 & 0.36 & 0.49 & 0.63 & 0.11 & 0.01 & 0.26 & 0.33 & 0.33 & 0.86 \\ 
    \bottomrule
    \end{tabular}
    \caption{PM represents $\prm$, for the other properties we represent their $\prm$ values, while Props represents the number of properties used for performing zero-shot testing}
    \label{tab:zero-shot}
\end{table}

\end{document}